\documentclass{article}

\usepackage{multirow}
\usepackage{diagbox}
\usepackage{algorithm}
\usepackage{algpseudocode}
\usepackage{graphicx}
\usepackage{amsmath}
\usepackage{amssymb}
\usepackage[T1]{fontenc}
\usepackage{neurips_2022}

\usepackage[utf8]{inputenc} 
\usepackage{hyperref}       
\usepackage{url}            
\usepackage{booktabs}       
\usepackage{amsfonts}       
\usepackage{nicefrac}       
\usepackage{microtype}      
\usepackage{xcolor}         
\usepackage{hyperref}

\title{Rethinking Reinforcement Learning based Logic Synthesis}

\author{%
  Chao Wang\thanks{Equal contributions} \\
  Huawei Noah's Ark Lab\\
  \texttt{w-c15@tsinghua.org.cn} \\
  \And
  Chen Chen\footnotemark[1] \\
  Huawei Noah's Ark Lab \\
  \texttt{chenchen9@huawei.com} \\
  \And
  Dong Li \\
  Huawei Noah's Ark Lab \\
  \texttt{lidong106@huawei.com} \\
  \And
  Bin Wang \\
  Huawei Noah's Ark Lab \\
  \texttt{wangbin158@huawei.com} \\
}

\begin{document}

\maketitle

\begin{abstract}
Recently, reinforcement learning has been used to address logic synthesis by formulating the operator sequence optimization problem as a Markov decision process. However, through extensive experiments, we find out that the learned policy makes decisions independent from the circuit features (i.e., states) and yields an operator sequence that is permutation invariant to some extent in terms of operators. 
Based on these findings, we develop a new RL-based method that can automatically recognize critical operators and generate common operator sequences generalizable to unseen circuits. Our algorithm is verified on both the EPFL benchmark, a private dataset and a circuit at industrial scale. Experimental results demonstrate that it achieves a good balance among delay, area and runtime, and is practical for industrial usage.
\end{abstract}

\section{Introduction}
Logic synthesis (LS) aims at finding an equivalent representation of large-scale integrated circuits \cite{mishchenko2007abc}.
It mainly consists of three steps, namely pre-mapping optimizations, technology mapping and post-mapping optimizations. In the pre-mapping optimization phase, technology-independent transformations are performed on the And-Inverter-Graph (AIG) representation of a circuit to reduce the graph's total number of nodes and levels\cite{ziegler2016synthesis}\cite{riener2019scalable}\cite{yu2018developing}\cite{haaswijk2018deep}\cite{hosny2020drills}\cite{zhu2020exploring} . In the technology mapping and post-mapping optimization phases, the AIG is mapped into standard ASIC or FPGA cells, followed by technology-dependent optimizations such as up-sizing and downsizing \cite{mishchenko2007abc}\cite{liu2017parallelized}.

Usually, a large portion of the
optimization efforts is on pre-mapping optimizations, which are carried out by a sequence of operators such as {\itshape rewrite}, {\itshape refactor}, and {\itshape resub} \cite{mishchenko2007abc}. Some of these operators are good at reducing the number of nodes of an AIG, and others the number of levels.
Pre-mapping optimization targets at finding an operator sequence that can reduce an AIG's total number of nodes (a.k.a., area), levels (a.k.a., delay) or both of them while keeping its boolean function unchanged. Since there are a large number of operators and the sequence length could be extremely long, combination of different operators and their permutations result in an exponentially growing search space. Additionally, each operator is associated with a set of tunable hype-parameters, making the LS problem even more difficult. For ease of discussion, as most of existing works do \cite{hosny2020drills,zhu2020exploring, himap2022DAC,boils}, we herein only consider the operator sequence optimization, setting each operator's hype-parameters as default. There are a number of pre-mapping optimization sequences, among which {\itshape Resyn2} is a well-known one.

As LS is a sequential decision making problem, some recent works formulate it as a Markov decision process (MDP) \cite{liu2017parallelized} and use reinforcement learning (RL) algorithms to attack it. RL learns a control policy that determines which operator to use at different states, where states are feature vectors of the current AIG. \cite{hosny2020drills} extracts some statistical features as the feature vector. In contrast, \cite{zhu2020exploring} and \cite{haaswijk2018deep} transform an AIG into a feature vector using graph neural networks (GNN). The policy is learned in a trial-and-error manner, maximizing s long-term accumulated reward, which is defined as a function of the AIG's reduced number of nodes or levels. Experiment results demonstrate their efficiency in terms of area or delay reduction. However, runtime of these algorithms is completely ignored. All of these methods train a different policy for different circuit online and then choose the best-performed one among tens or hundreds of online evaluation trials. The training and evaluation procedures often take hundreds of or thousands of times the runtime cost of {\itshape Resyn2}, which is totally unacceptable for real applications.

Based on these works, we conduct extensive experiments and have two basic findings: (1) decisions made by the RL policy do not depend on circuit features; (2) permutation of these operators have little effect on the final performance. These findings reveal that: 1) it is unnecessary to extract circuit features 2) though LS has an exponentially growing search space, the loss surface is somewhat flat, since permutation of operators in the same sequence has little effect on the final performance.

Based on these findings, we develop a new method that can automatically recognize critical operators and generate a common sequence for different circuits. The method consists of two major steps, where the first step learns a shared policy for a number of circuits (served as a training dataset), and the second step searches for a best-performed common sequence based on the learned policy. Different from previous RL-based methods that recommend a different operator sequence for different circuit, the common sequence can be directly used to optimize unseen circuits without online learning or further adaptation, saving lots of time. Extensive experiment results demonstrate that our method is capable of finding a common sequence that achieves good performance in terms of area or delay reduction and significantly reduce the runtime. The common sequence can be directly integrated into ABC \cite{mishchenko2007abc}, a well-known LS tool, as a substitute for {\itshape Resyn2}, without modifying the ABC architecture.

The remainder of this paper is organized as follows. Section \ref{sec:background} gives a brief introduction to LS and RL. Section \ref{sec:rethinking} elaborates our basic findings on RL based LS. Section \ref{sec:method} introduces our proposed method. Section \ref{sec:experiment} presents experiment results and section \ref{conclusion} concludes this paper.

\section{Background}\label{sec:background}
\subsection{Logic Synthesis}
The whole process of LS can be divided into three stages, i.e., pre-mapping optimizations, technology mapping and post-mapping optimizations. The pre-mapping optimization stage is at the core of the LS process. There exist a rich set of optimization operators, such as {\itshape rewrite}, {\itshape refactor}, and {\itshape resub}. Some of these operators are good at reducing the number of nodes of an AIG, and others the number of levels.

An AIG often has thousands of or even millions of nodes, and the number of levels can also be large. Most of these operators optimize an AIG by iteratively replacing the subgraph rooted at a chosen node with a new subgraph that has less number of nodes or levels, where the root node is selected in a topological order. In this way, the whole AIG could be optimized \cite{mishchenko2007abc}. As some operators are good at reducing area and some delay, it is necessary to design an operator sequence that can make full use of different operators. As the length of an operator sequence could be large, the search space grows exponentially with the sequence length.

\subsection{Reinforcement Leaning}
MDPs \cite{sutton2018reinforcement} are widely used to formulate discrete time stochastic control processes. 
Consider a MDP defined by a tuple $(\mathcal{M}, \mathit{S}, \mathit{A}, r, \mathit{P},\gamma, H)$, where $\mathit{S}$ is the state space, $\mathit{A}$ is the
finite action space, $r:\mathit{S} \times \mathit{A} \rightarrow \mathbb{R}$ the bounded reward function, $\mathit{P}:S \times A \times S \rightarrow \mathbb{R}_{\in [0,1]}$ is the transition probability distribution, $\gamma$ is the discount factor that we assume $\gamma \in [0,1)$ and $H$ is the episode horizon.
The agent interacts with the MDP at discrete time steps by performing its policy $\pi: S \times A \rightarrow \mathbb{R}_{\in [0,1]}$, 
generating a trajectory of states and actions,
$\tau = \left(s_0, a_0, s_1, a_1, \dots s_H, a_H \right)$, where $s_0 \sim \rho_0(s)$, $a_t \sim \pi(\cdot | s_t)$ and $s_{t+1} \sim \mathit{P}(\cdot |s_t, a_t)$.

The objective of a reinforcement learning agent is to 
find a policy that 
maximizes the expected cumulative discounted rewards given the initial state$s_0$:
\begin{equation}
\label{eqation:RL-objective}
    \mathit{J}(\pi) = \mathbb{E}_{ a_t \sim \pi(\cdot|s_t),s_{t+1}\sim P(\cdot|s_t,a_t)} \left[\sum_{t=0}^{H}\gamma^{t} r(s_t, a_t) \right].
\end{equation}

REINFORCE is a RL algorithm that parameterizes the policy $\pi(a|s)$ as a continuous function $\pi(a|s,\theta)$ with $\theta$ as the function parameters. $\theta$ is updated by the policy gradient method \cite{sutton2000policy}:
\begin{equation}
 \theta = \theta + \alpha \nabla_{\theta}log \pi_{\theta}(s_t,a_t) \sum_{i=t}^{H}r_i, \end{equation}
where $\alpha$ is the learning rate.

\subsection{Problem Formulation}\label{sec:problem_formulation}
As done in \cite{liu2017parallelized}\cite{hosny2020drills}\cite{zhu2020exploring}, given a circuit, LS can be formulated as a MDP, where the state space $S$ contains all equivalent AIGs and the action space $A$ contains available operators. In this paper, we fix the action space as $A = \{resub, resub -z, rewrite, rewrite -z,refactor, refactor -z, balance \}$, which is the same as that used by {\itshape Resyn2}. The initial state $s_0$ is the original AIG to be optimized. Given any $s \in S$  and $a \in A$, $s'$ is an AIG obtained by applying an operator $a$ to AIG $s$.  The reward function $r(s,a,s')$ can be designed as a function of the reduced amount of area or delay after applying the operator $a$ (e.g., the weighted summation of the reduced area or delay). Then we can leverage RL algorithms to learn a policy that can transform a circuit into a simpler one.

Different reward functions are designed in \cite{liu2017parallelized}\cite{hosny2020drills}\cite{zhu2020exploring}, some encourage area reduction under a delay constraint, some encourage delay reduction, etc. For ease of discussion, in this paper, we employ two simple reward schemes, one encourages area reduction and the other delay reduction, denoted as area-first and delay-first reward schemes, respectively. Specifically, for the area-first reward scheme, $r(s,a,s')$ denotes the ratio of the reduced area of the two AIGs $s$ and $s'$ over the initial area. Similarly, for the delay-first reward scheme, $r(s,a,s')$ is the ratio reduced delay over the initial delay.

\section{Revisiting RL-based Logic Synthesis}\label{sec:rethinking}

In this section, we dive deep into current RL-based LS methods. Specifically, we first evaluate the influence of different feature embeddings of an AIG on the LS performance, and then investigate the behavior of the learned policy during the decision-making phase. We select four circuits in the EPFL benchmark \cite{amaru2015epfl} and four from \cite{mishchenko2007abc},\cite{brglez1989combinational} and \cite{yang1991logic} as a circuit dataset, denoted as EPFL-Test. Besides, we use REINFORCE as the learning algorithm and follow similar hyper-parameter configurations as done in \cite{zhu2020exploring}.

\begin{table*}[h]
    \caption{LS Results of different RL-based algorithms with different feature embeddings. Each number in the Area and Delay column denotes the true value of the number of nodes or levels. Those in the "Imp." columns are the ratios of the reduced area or delay over their initial values.}
    \label{tab:action_distribution}
    \begin{center}
     \scalebox{0.73}{
        \begin{tabular}{c|c|c|c|c|c|c|c|c|c|c|c}
            \hline
            \hline
            \multicolumn{2}{c}{\multirow{2}{*}{\diagbox{Dataset}{Results}}} & \multicolumn{2}{|c|}{Init.} & \multicolumn{2}{c|}{Resyn2} & \multicolumn{3}{c|}{Area-first} & \multicolumn{3}{c}{Delay-first} \\
            \cline{3-12}
            & &Area & Delay & Area Imp. & Delay Imp. & Area & Delay & Area Imp. & Area & Delay & Delay Imp.\\
            \hline
            \multirow{9}{*}{Graph} 
            & log2 & 32060 & 444 & 0.08 & 0.15 & 29788 & 385 & 0.07 & 31467 & 398 & 0.1 \\
            \cline{2-12}
            & multi. & 27062 & 274 & 0.09 & 0.04 & 24776 & 274 & 0.08 & 26888 & 266 & 0.03 \\
            \cline{2-12}
            & sin & 5416 & 225 & 0.07 & 0.21 & 5066 & 179 & 0.07 & 5267 & 203 & 0.097 \\
            \cline{2-12}
            & sqrt & 24618 & 5088 & 0.21 & 0.02 & 18995 & 5076 & 0.23 & 19552 & 5033 & 0.01 \\
            \cline{2-12}
            & c1355 & 502 & 25 & 0.22 & 0.32 & 390 & 18 & 0.22 & 397 & 18 & 0.28 \\
            \cline{2-12}
            & c5315 & 1776 & 37 & 0.29 & 0.27 & 1523 & 36 & 0.14 & 1388 & 27 & 0.27 \\
            \cline{2-12}
            & c7552 & 1469 & 26 & 0.03 & 0.0 & 1400 & 26 & 0.047 & 1466 & 26 & 0 \\
            \cline{2-12}
            & i10 & 2675 & 50 & 0.31 & 0.36 & 1788 & 44 & 0.33 & 1055 & 33 & 0.34 \\
            \cline{2-12}
            & AVERAGE & - & - & 0.1625 & 0.17 & - & - & 0.1483 & - & - & 0.133 \\
            \hline
            \hline
            \multirow{9}{*}{Stat.} 
            & log2 & 32060 & 444 & 0.08 & 0.15 & 29730 & 390 & 0.016 & 31545 & 382 & 0.14 \\
            \cline{2-12}
            & multi. & 27062 & 274 & 0.09 & 0.04 & 24627 & 263 & 0.09 & 26853 & 274 & 0.0\\
            \cline{2-12}
            & sin & 5416 & 225 & 0.07 & 0.21 & 5076 & 181 & 0.06 & 5330 & 201 & 0.10 \\
            \cline{2-12}
            & sqrt & 24618 & 5088 & 0.21 & 0.02 & 19318 & 5031 & 0.22 & 19439 & 5021 & 0.01 \\
            \cline{2-12}
            & c1355 & 502 & 25  & 0.22 & 0.32 & 390 & 18 & 0.22 & 389 & 18 & 0.28 \\
            \cline{2-12}
            & c5315 & 1776 & 37 & 0.29 & 0.27 & 1643 & 37 & 0.07 & 1361 & 28 & 0.24 \\
            \cline{2-12}
            & c7552 & 1469 & 26 & 0.03 & 0.0 & 1447 & 26 & 0.01 & 1469 & 26 & 0.0 \\
            \cline{2-12}
            & i10 & 2675 & 50 & 0.31 & 0.36 & 1787 & 43 & 0.33 & 1833 & 33 & 0.34 \\
            \cline{2-12}
            & AVERAGE & - & - & 0.1625 & 0.17 & - & - & 0.127 & - & - & 0.139 \\
            \hline
            \hline
            \multirow{9}{*}{Rand.} 
            & log2 & 32060 & 444 & 0.08 & 0.15 & 29299 & 387 & 0.09 & 30186 & 377 & 0.15 \\
            \cline{2-12}
            & multi. & 27062 & 274 & 0.09 & 0.04 & 24388 & 263 & 0.1 & 25079 & 262 & 0.04 \\
            \cline{2-12}
            & sin & 5416 & 225 & 0.07 & 0.21 & 5053 & 171 & 0.07 & 5066 & 171 & 0.24\\
            \cline{2-12}
            & sqrt & 24618 & 5088 & 0.21 & 0.02 & 19250 & 5058 & 0.22 & 19331 & 4971 & 0.02 \\
            \cline{2-12}
            & c1355 &  502 & 25 & 0.22 & 0.32 & 392 & 17 & 0.22 & 392 & 17 & 0.32 \\
            \cline{2-12}
            & c5315 & 1776 & 37 & 0.29 & 0.27 & 1368 & 32 & 0.23 & 1362 & 28 & 0.24 \\
            \cline{2-12}
            & c7552 &1469 & 26 & 0.03 & 0.0 & 1394 & 26 & 0.05 & 1400 & 26 & 0 \\
            \cline{2-12}
            & i10 & 2675 & 50 & 0.31 & 0.36 & 1782 & 36 & 0.33 & 1796 & 33 & 0.34 \\
            \cline{2-12}
            & AVERAGE & - & - & 0.1625 & 0.17 & - & - & 0.1637& - & - & 0.16875 \\
            \hline
        \end{tabular}
        }
    \end{center}
\end{table*}

\subsection{Circuit Features}\label{subsec:state_representation}
Since the state is an AIG with thousands of or even millions of nodes and edges, feature extraction is needed to transform the raw AIG into a compact representation. Existing representation methods  can be roughly divided into two categories. \\
$\bullet$  \textbf{Statistics features} 
\cite{hosny2020drills} use the statistical information of an AIG, [$\# primary I/O$, $\# nodes$,
$\# edges$,$\# levels$,$\# latches$,$\% ANDs$,$\% NOTs$], to represent the circuit. When an operator is applied, these statistics will change. Therefore, statistics embedding indeed includes some important information of the circuit.  \\
$\bullet$  \textbf{Graph features}
\cite{haaswijk2018deep}\cite{zhu2020exploring} propose to use not only the statistical features but also the graph embedding of an AIG. To obtain the graph embedding, each node in the AIG is firstly represented as a vector (node embedding) containing information about its node type, as well as the edge type of its two fanins. Node embeddings are then passed through two consecutive layers of graph convolution networks so that each node's neighborhood information is aggregated. The graph features is then obtained by taking the average of all the node embeddings. 

We argue that both the statistics feature and the graph feature are incapable of extracting useful features. Firstly, it is obvious that the statistics feature loses too much structural information of an AIG, and it is quite possible that two thoroughly different AIGs have the same statistical features. 
Secondly, the current the graph feature extracting scheme is virtually ineffective and impractical. The reasons are two fold. One on hand, aggregating millions of node features into a single feature vector is still open to discussion, and computing the graph embedding requires huge computational overhead. On the other hand, most LS operators are in fact local optimization operators that iteratively choose one node each time in an AIG and replace the subgraph rooted at the chosen node (a.k.a., cut \cite{mishchenko2007abc}) with a new subgraph that has the same boolean function but less nodes or levels. In principle, aggregating all the nodes information into a single vector can not guide operators to conduct subgraph replacements.

Our empirical studies also verify the above claims. We perform RL-based LS using different types of circuit features as circuit states, i.e., statistics features, graph features and non-informative features (random vectors having the same dimension as that of statistics features). For ease of denotation, we name non-informative features as random features. Experiment results are shown in Table \ref{tab:action_distribution}. As we can see, the area and delay given by different RL algorithms are reduced by around 16\%, which are similar to those of {\itshape Resyn2}. Furthermore, it seems that these RL policies can make decisions without accessing AIG states, since the policy performs well even using random features.


\subsection{Permutation Invariance}\label{subsec:permutation}
\begin{figure}
    \centering
    \includegraphics[width=14cm]{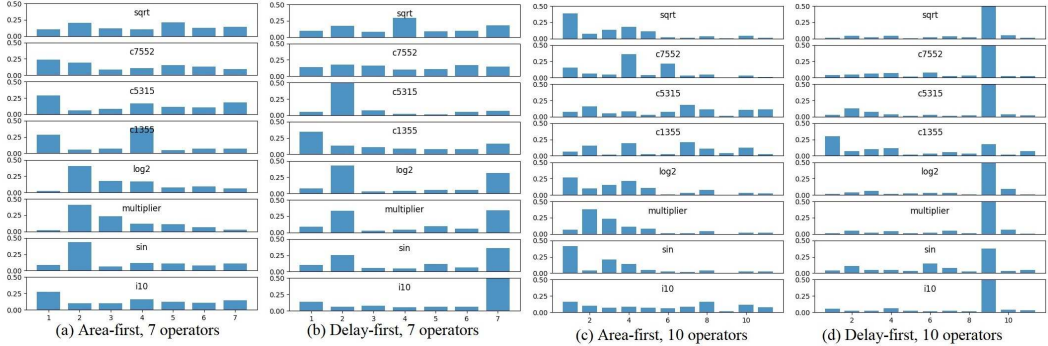}
    \caption{Operator distributions of different circuits. Note that the maximum value of the y-axis is clipped to 0.5.}
    \label{fig:operator_distribution}
\end{figure}

Since decisions are made without accessing circuit features (i.e., $\pi(a|s)$ degenerates to $\pi(a)$), we guess the action distributions at different time steps are similar.
To this end, we investigate the decisions made by them at different time steps. As expected, the learned policy samples an operator from an operator distribution that hardly evolves over time, where operator distributions for different circuits are illustrated in Fig \ref{fig:operator_distribution} (due to space limitation, we only show the operator distributions of the policy using random feature features. Policies using statistic features and graph features have similar results. Besides, we also show the learned operator distributions of the RL algorithm with an extended operator space, which we would discuss later).

The above finding suggests that operators may be permutation invariant to some extent. To verify it, for each circuit, we first sample an operator sequence (consisting of ten operators) according to the operator distribution shown in Fig \ref{fig:operator_distribution}, and then generate ten sequences by randomly shuffling the sampled sequence. Evaluation results of these randomly permuted sequences are shown in Table \ref{tab:operator_permutation}. As expected, there is only a minor difference among the results of these sequences, demonstrating that the performance of an operator sequence is almost permutation invariant. Based on these findings, we can 
conclude that although LS has an exponentially growing search space, the loss surface seems relatively flat and the (local) optima is easy to reach.

\subsection{Critical Operators}\label{subsec:critical_operators}
From Fig \ref{fig:operator_distribution}, we can see that the performance improvements are mainly determined by some frequently used operators, which we denote as critical operators, and RL-based algorithms could learn to discover these critical operators. Then we wonder if the performance would be further improved by involving new operators. In this regard, we re-implement the above RL algorithms with an extended operator space (newly added operators are {\itshape sopb, \itshape fraig, \itshape dsdb, \itshape blut}). As shown in Table \ref{tab:more_operators}, when more operators are used, the area remains similar but the delay is reduced from 0.82 to 0.52 (these numbers are the mean of the ratios of the area or delay of the optimized circuit over the those of the initial circuit, and the smaller, the better). The results make sense since most of the four newly added operators are good at delay optimization.

These results demonstrate that the newly incorporated operators do expand the optimization upper bound of LS and RL algorithms are still capable of discovering good operators suitable for different circuits.

\begin{table}[h]
    \caption{RL-based LS with more Operators. Each number in the table is the mean of ratios of the optimized area or delay over their initial values (the smaller, the better).}
    \label{tab:more_operators}
    \begin{center}
        \begin{tabular}{c|c|c|c|c|c|c|c}
            \hline
            \hline
            \multicolumn{2}{c|}{Init.} & \multicolumn{2}{c|}{Resyn2} & \multicolumn{2}{c|}{Area-first} & \multicolumn{2}{c}{Delay-first} \\
            \hline
            Area & Delay & Area & Delay &  Area & Delay & Area & Delay \\
            \hline
            1.0 & 1.0 & 0.83 & 0.82 & 0.83 & 0.84 & 1.009 & 0.52 \\
            \hline
        \end{tabular}
    \end{center}
\end{table}

\begin{table}[h]
    \caption{Permutation Invariance of Operator Sequences. Ten randomly permutated sequences are evaluated for each circuit. Each number in the "Mean", "Max" and "Min" column denotes the area or delay of an AIG normalized by its initial values (the smaller, the better). }
    \label{tab:operator_permutation}
    \begin{center}
    
        \begin{tabular}{c|c|c|c|c|c|c|c|c}
            \hline
            \hline
            \multirow{2}{*}{Name} 
            & \multicolumn{4}{c|}{Area-first} & \multicolumn{4}{c}{Delay-first} \\
            \cline{2-9}
            & Mean & Max & Min & Std & Mean & Max & Min & Std\\
            
            \hline
            log2 & 0.92 & 0.93 & 0.91 & 0.005 & 0.93 & 0.94 & 0.86 & 0.012 \\
            \hline
            multi. & 0.90 & 0.91 & 0.89 & 0.003 & 0.11 & 0.93 & 0.96 & 0.01 \\
            \hline
            sin & 0.93 & 0.94 & 0.93 & 0.003 & 0.93 & 0.95 & 0.79 & 0.007 \\
            \hline
            sqrt & 0.79 & 0.80 & 0.78 & 0.006 & 0.78 & 0.79 & 0.99 & 0.003 \\
            \hline
            c1355 & 0.77 & 0.78 & 0.76 & 0.004 & 0.78 & 0.85 & 0.77 & 0.019\\
            \hline
            c5315 & 0.73 & 0.76 & 0.70 & 0.019 & 0.75 & 0.77 & 0.72 & 0.012 \\
            \hline
            c7552 & 0.94 & 0.95 & 0.94 & 0.004 & 0.95 & 0.96 & 0.93 & 0.006\\
            \hline
            i10 & 0.68 & 0.72 & 0.64 & 0.02 & 0.71 & 0.74 & 0.67 & 0.024\\
            \hline
        \end{tabular}
    \end{center}
\end{table}

In summary, we first find out that decisions made by RL policies do not depend on circuit features (states). The phenomenon in Subsection \ref{subsec:permutation} that the operator distribution does not evolve over time also verifies the state-agnostic property. Furthermore, it also indicates that operators are permutation invariant to some extent, which is supported by our experiments. Besides, we also observe that the performance could be further improved by incorporating new operators and RL is still capable of discovering critical operators.

\section{Method}\label{sec:method}

Though existing RL methods can learn policies with good performance, runtime is thoroughly ignored, which is of critical importance for real applications. Existing works \cite{hosny2020drills}\cite{zhu2020exploring} propose to learn different policies for different circuits. 
For each circuit, time elapsed for the feature extraction process and RL training process is often hundreds of thousands of times that of {\itshape Resyn2}. Therefore, when there are a number of circuits to be optimized, the time required is totally unacceptable. In this section, we propose a method that learns a common policy for different circuits using RL, based on which a low-runtime common operator sequence is derived.

\begin{algorithm}
\caption{RL-driven Common Sequence Design}
\label{alg:algorithm}
\begin{algorithmic}[1]
\Procedure{RL TRAINING}{$\mathcal{D}$, $T$}\quad $\triangleright$ 
RL training phase takes a circuit dataset $\mathcal{D}$ and the maximum training $T$ step as inputs
\State Randomly Initialize a policy function $\pi(a|s_r)$, where $s_r$ is sampled from a 4D uniform distribution each time $\pi(a|s_r)$ is used
    \For {$i=0, \cdots, T $}
        \State Randomly sample a circuit from the training set $\mathcal{D}$.
        \State Rollout an operator sequence following policy $\pi(a|s_r)$
        \State Update policy parameters using REINFORCE \cite{sutton2018reinforcement}
	\EndFor
	\State Return $\pi(a|s_r)$
\EndProcedure

\Procedure{SEQUENCE DESIGN}{$\mathcal{D}$, $\pi(a|s_r)$, $L$}\quad $\triangleright$  Sequence design phase takes the training set $\mathcal{D}$, the learned policy $\pi(a|s_r)$ and the sequence length $L$ as inputs 
    \State Initialize an empty sequence $Seq$ and a performance metric $m=0$
    \For {$i=0, \cdots, 9$}
        \State Sample a common sequence $S_i$ with length $L$ by following the learned policy $\pi(a|s_r)$
        \State Obtain the value of reduced area or delay $m_i$ of the common sequence by evaluating it on the training set $\mathcal{D}$
        \State Let $m = m_i$ and $Seq=Seq_i$ if $m_i > m$
	\EndFor
	\State Return $Seq$
\EndProcedure
\end{algorithmic}
\end{algorithm}

The reason of choosing RL is as follows. According to our analysis in Section \ref{sec:rethinking}, we can reach a conclusion that LS is a MDP with a really special structure: given a circuit, the optimal action distribution at different time steps hardly evolve over time. This is analogous to a barrier-free 2-dimensional navigation task. It is obvious that a talented agent should move towards the target position at each time step, and superficially, navigation decisions are made without accessing any states. For LS, an agent aims to push an initial AIG to another point (in the AIG space) that has less number of nodes or levels. Then executing an operator sampled from a fixed operator distribution at different time steps is analogous to executing a fixed action in the 2D navigation task. Accordingly, we continue to leverage RL algorithms with random state representations to learn policies, as done in Subsection \ref{subsec:state_representation}. Using random features saves plenty of time especially when compared with the approach using graph features.

Furthermore, we need to learn a policy that can generalize to unseen circuits. Only in this way, we can avoid online learning a new policy for a new circuit, making RL a feasible approach for real applications. Fortunately, the operator distributions are similar to each other, as shown in Fig \ref{fig:operator_distribution}. Consequently, we expect a common distribution would work for different circuits. To this end, we no longer train a separate policy for each circuit. Instead, we train a shared policy on a training dataset consisting of multiple circuits. During the training phase, the algorithm randomly samples a circuit from the training dataset, rollouts an operator sequence using the currently learned policy and then performs policy gradient update using the collected data.

Additionally, observing that operators are somewhat permutation invariant, we further derive a common sequence based on the learned common policy. Specifically, we randomly sample 10 sequences (each composed of 10 operators) according to the learned policy and select the best-performed one based on their evaluation results on the training dataset. The selected common sequence can serve as an alternative to {\itshape Resyn2}, without modifying the structure of the ABC LS tool.

\begin{table*}[h]
    \caption{Results of different RL-based algorithms on the six unseen circuits from the EPFL benchmark. The 'Common' row is the results of our common sequence. Runtime is normalized by the elapsed time of the common sequence. Each number in the area and delay columns is the mean of ratios of the area or delay over their initial values (the smaller, the better).}
    \label{tab:common_sequence_results}
    \begin{center}
        \scalebox{0.98}{\begin{tabular}{c|c|c|c|c|c|c|c|c}
            \hline
            \hline
            \multirow{2}{*}{\diagbox{Methods}{Results}} & \multicolumn{2}{|c|}{Init.}  & \multicolumn{3}{c|}{Area-first} & \multicolumn{3}{c}{Delay-first} \\
            \cline{2-9}
            &Area & Delay & Area & Delay & Runtime & Area & Delay & Runtime \\
            \hline
            Graph embeddings & 1.0 & 1.0 & 0.89 & 0.93 & 1000+ & 10.921 & 0.943 & 1000+ \\
            \hline
            Stats. embeddings & 1.0 & 1.0 & 0.91 & 0.992 & 200+ & 0.9355 & 0.999 & 200+  \\
            \hline
            Common sequence & 1.0 & 1.0 & 0.876 & 0.94 & 1.0 & 0.845 & 0.95 & 1.0 \\
            \hline
        \end{tabular}}
    \end{center}
\end{table*}
In summary, our proposed method first learns a common policy on a training dataset consisting of multiple circuits through RL, and then derives a common operator sequence based on the learned policy. The common sequence can be directly used for new circuits without online learning or further adaptation. Our proposed method is demonstrated in algorithm \ref{alg:algorithm}.

\section{Experiments}\label{sec:experiment}
In this section we present the main results of our proposed method. We first evaluate its runtime efficiency and then its generalization ability to unseen circuits. The maximum training epochs $T$ in algorithm \ref{alg:algorithm} is set to 200. Besides, in Section \ref{sec:generalization} and \ref{sec:scale_up}, to meet with industrial demands, we extend the fixed operator set introduced in Section \ref{sec:problem_formulation} by adding the four operators {\itshape sopb, \itshape fraig, \itshape dsdb, \itshape blut} used in Section \ref{sec:generalization}.

\subsection{Learning a Common Sequence}
We feed the EPFL-Test dataset into algorithm \ref{alg:algorithm} and it returns a common operator sequence, which is directly evaluated on the left six circuits from the EPFL benchmarks \cite{amaru2015epfl}. Besides, the two RL algorithms using graph features and statistics features are re-implemented to learn a new policy online for each of the six circuits, as done in \cite{hosny2020drills} and \cite{zhu2020exploring} (each policy is also trained for 200 epochs). Experiment results are shown in Table \ref{tab:common_sequence_results}. As we can see, these RL-based online learning algorithms and our common sequence yield similar results, demonstrating that our RL-based common sequence design method is indeed capable of automatically generating common sequences that perform well on new circuits. 

Most importantly, since our method does not require online learning a new policy or searching a new sequence for an unseen circuit, its runtime cost is comparable with that of {\itshape Resyn2}. On the contrary, the RL algorithm using statistics features needs to learn a new policy for each unseen circuit, which takes more than two hundred times the runtime of ours. Worse still, the RL algorithm using graph features needs to extract graph features and the runtime cost is even worse. The online learning paradigm of existing RL solutions prevents them from real applications. 

\subsection{Generalization to New Circuits}\label{sec:generalization}

\begin{table*}[h]
    \caption{Generalization to new circuits. Each number in the area and delay columns is the average of the ratios of the optimized area or delay over their initial values (the smaller, the better). Runtime is normalized by that of {\itshape Resyn2.}}
    \label{tab:generalization_to_new_circuits}
    \begin{center}
        \scalebox{0.79}{\begin{tabular}{c|c|c|c|c|c|c|c|c|c|c|c}
        
           \hline\hline
           \multirow{2}{*}{\diagbox{Dataset}{Results}} & \multicolumn{2}{|c|}{Init.} & \multicolumn{3}{c|}{Resyn2} & \multicolumn{3}{c|}{Area-first} & \multicolumn{3}{c}{Delay-first} \\
            \cline{2-12}
            & Area & Delay & Area & Delay & Runtime & Area & Delay & Runtime & Area & Delay & Runtime \\
            \hline
            Training & 1.0 & 1.0 & 0.72 & 0.70 & 1.0 & 0.72 & 0.68 & 1000+ & 0.77 & 0.47 & 1000+  \\
            \hline
            Evaluation & 1.0 & 1.0 & 0.80 & 0.87 & 1.0 & 0.77 & 0.86 & 3.7 & 0.92 & 0.59 & 5.6  \\
            \hline
        \end{tabular}}
    \end{center}
\end{table*}
To evaluate the generalization ability of our method, we introduce a private dataset (denoted as \textcolor{black}{INDU}) that \textcolor{black}{was originally for industrial uses} and consists of 40 circuits. We select ten of them as the training set and evaluate the common sequence on the rest. Results are shown in Table \ref{tab:generalization_to_new_circuits}. 

As observed, the common sequence significantly reduces the area (area-first) and the delay (delay-first) on the training set to 0.68 and 0.47, respectively, and the two values are reduced to 0.77 and 0.59 on the evaluation set. Although the common sequence is trained on 10 circuits, it generalizes well to the left 30 circuits that do not appear in the training set. Besides, the common sequence only consists of 10 operators and is consequently runtime efficient. 

\begin{table*}[h]
    \centering
    \caption{LS results on a large-scale circuit C347. Note that each number in the last row denotes the area and delay of the circuit \emph{after technology mapping}.}
    
    \scalebox{1}{\begin{tabular}{c|c|c|c|c|c|c|c|c}
    \hline
    \hline
         \multirow{2}{*}{Circuit} 
         & \multicolumn{2}{c}{Init.} & \multicolumn{3}{|c}{Resyn2} & \multicolumn{3}{|c}{Common Sequence} \\
         \cline{2-9}
         & Area & Delay & Area & Delay & Runtime & Area & Delay & Runtime \\
         \hline
         C347  & 615886 & 223 & 365002 & 195 & 414s & 344040 & 176 & 1655s \\ 
         \hline
    \end{tabular}}
    
    \label{tab:vmac190}
\end{table*}

\subsection{Evaluation on a Industrial Circuit}\label{sec:scale_up}
To meet with the \textcolor{black}{industrial demand}, we design a new common sequence by re-implementing our method on the entire INDU benchmark with slightly different configurations. Specifically, our previous experiments use the reduced number of nodes or levels \emph{before technology mapping} as the reward function. Here we shift the reward function to the reduced amount of area or delay of an AIG \emph{after technology mapping}. The learned common sequence (consisting of 10 operators) is then used to optimize a large-scale circuit named \textcolor{black}{C347} with around 3,470,000 nodes and 700 levels. Results are shown in Table \ref{tab:vmac190}. Again, our method generates a common sequence that shows impressive performance on the circuit which is significantly different from the training set in terms of circuit scale (i.e., the number of nodes and levels). 


\section{Conclusion}\label{conclusion}
In this paper, we propose a practical RL-based method that can automatically recognize critical operators and generate a common sequence for LS tasks. Specifically, we first investigate current RL-based methods and find out that the learned policy is state-agnostic and yields an operator sequence that is somewhat permutation invariant. In this regard, we propose to learn a common policy using RL and then derives a common sequence on the training set consisting of multiple circuits. Experiment results demonstrate that the common sequence can reduce the area or delay of different circuits and can be directly applied to new circuits without online learning or further adaptation. Furthermore, when more operators are used, our method is still capable of finding good operator distributions and generating common sequences that reduce the area and delay by a large margin.
Last but not least, our method is runtime-efficient when compared with existing RL-based methods that need to learn a new policy for a new circuit. Besides, the designed common sequence can serve as a substitute for {\itshape Resyn2} baseline without modifying the architecture of ABC, the open-source LS tool. In future work, we will continue exploring novel sequence and parameter optimization methods alone the line that does not access circuit features to better solve the LS problem. 

\bibliographystyle{plainnat}
\bibliography{citation}
\appendix



\end{document}